%% file: main.tex
\documentclass[a4paper]{article}

\usepackage{INTERSPEECH2020}
\usepackage{amsmath}
\usepackage{amssymb}
\usepackage{booktabs}
\usepackage{multirow}
\usepackage{hyperref}
\usepackage{enumitem}
\hypersetup{
    colorlinks=true,
    citecolor=blue,
    linkcolor=blue,
    urlcolor=blue,
}

\title{SpeechStew: Simply Mix All Available Speech Recognition Data to \\ Train One Large Neural Network}
\name{William Chan, Daniel S. Park, Chris A. Lee, Yu Zhang, Quoc V. Le, Mohammad Norouzi}
\address{
  Google Research, Brain Team}
\email{\{williamchan,danielspark,chrisalee,ngyuzh,qvl,mnorouzi\}@google.com}

\begin{document}
\maketitle
\begin{abstract}
We present SpeechStew, a speech recognition model that is trained on a combination of various publicly available speech recognition datasets: AMI, Broadcast News, Common Voice, LibriSpeech, Switchboard/Fisher, Tedlium, and Wall Street Journal. SpeechStew simply mixes all of these datasets together, without any special re-weighting or re-balancing of the datasets. SpeechStew achieves SoTA or near SoTA results across a variety of tasks, without the use of an external language model. Our results include 9.0\% WER on AMI-IHM, 4.7\% WER on Switchboard, 8.3\% WER on CallHome, and 1.3\% on WSJ, which significantly outperforms prior work with strong external language models. We also demonstrate that SpeechStew learns powerful transfer learning representations. We fine-tune SpeechStew on a noisy low resource speech dataset, CHiME-6. We achieve 38.9\% WER without a language model, which compares to 38.6\% WER to a strong HMM baseline with a language model.
\end{abstract}


\section{Introduction}
End-to-end speech recognition models \cite{graves-icml-2014,chan-icassp-2016,chan-icml-2020} have seen remarkable success in recent years~\cite{park-interspeech-2019}. For instance, end-to-end speech recognition models have achieved state-of-the-art (SoTA) results on LibriSpeech \cite{zhang-arxiv-2020} and large proprietary datasets \cite{chiu-icassp-2018}. The success of these methods have  often been attributed to the abundance of training data \cite{chiu-icassp-2018} and the use of large deep models \cite{zhang-arxiv-2020}. However, on noisy, low resource speech recognition datasets, such as CHiME-6 \cite{watanabe-arxiv-2020}, where overfitting is a significant problem, end-to-end methods tend to struggle relative to HMM-based baselines~\cite{andrusenko-arxiv-2020}. For example, the best previously published end-to-end model achieved 49.0\% WER on the CHiME-6 dev set \cite{andrusenko-arxiv-2020}, while the best HMM model achieves 36.9\% WER \cite{medennikov-chime-2020}.

Multi-lingual training \cite{burget-icassp-2010,heigold-icassp-2013}, multi-domain training \cite{kaldi-multi-en,narayanan-slt-2018}, unsupervised pre-training \cite{devlin-naacl-2019,baevski-arxiv-2020}, semi-supervised learning \cite{synnaeve-icml-2020,park-interspeech-2020} and transfer learning \cite{donahue-icml-2014,kornblith-cvpr-2019} are some techniques proposed in the literature to enhance generalization. These methods optimize speech recognition models on data from related tasks (typically of high resource), to help the specific task of interest (typically of low resource). For example, in multi-lingual training, the knowledge from a high resource language may transfer to a low resource language \cite{li-icassp-2019}. In multi-domain training, combining different domain datasets of the same language, could facilitate the cross-sharing of knowledge across the domains \cite{narayanan-slt-2018}. In unsupervised pre-training, the knowledge from the pre-training task may transfer to the supervised task~\cite{schneider-arxiv-2019}. In transfer learning, a general model is trained with a large amount of data. Subsequently, its knowledge is transferred via fine-tuning on training data from a downstream task that is typically of low resource~\cite{kolesnikov-arxiv-2019}.

Neural networks have benefited from simply scaling up to larger and deeper models across multiple application domains like computer vision~\cite{krizhevsky-nips-2012,he-cvpr-2016} and language modelling \cite{vaswani-nips-2017,brown-neurips-2020}. It has been observed that these large models tend to enhance generalization \cite{nakkiran-iclr-2020}. End-to-end speech recognition models have also been benefited from the increases in model capacity \cite{zhang-icassp-2017,gulati-interspeech-2020,zhang-arxiv-2020}.

This paper presents SpeechStew. SpeechStew is a simple approach to end-to-end speech recognition, which leverages both multi-domain training and transfer learning. 
SpeechStew follows the following simple recipe:
\begin{enumerate}[noitemsep]    
    \item Combine all available speech recognition data without any domain-dependent re-balancing or re-weighting.
    \item Train a single large neural network (a 100M or 1B parameter model) on the combined data.
\end{enumerate}
Our method does not utilize any domain labels, or introduce any additional hyperparameters for combining the data. We do not incorporate an external language model during inference, yet our result compares favourably to prior work that utilize strong language models, achieving SoTA or near SoTA results across various tasks (AMI, Common Voice, LibriSpeech, Switchboard, Tedlium, and WSJ).

We also demonstrate that SpeechStew has strong transfer learning capabilities. When presented with a new unseen low resource dataset (CHiME-6 in our setup), we merely:

\begin{enumerate}[noitemsep]
    \setcounter{enumi}{2}
    \item Fine-tune SpeechStew on the new labelled dataset.
\end{enumerate}

We find that this straightforward pre-training and fine-tuning procedure yields near-SoTA results on CHiME-6. This is encouraging since CHiME-6 is a particularly challenging task \cite{watanabe-arxiv-2020} for end-to-end speech recognition models, which suffer from over-fitting issues \cite{andrusenko-arxiv-2020}. We also demonstrate that our method is complementary to other pre-training methods, in particular unsupervised wav2vec pre-training \cite{schneider-arxiv-2019} which we use in conjunction with SpeechStew training.

The key contributions of this paper are as follows:
\begin{enumerate}[noitemsep]
    \item We demonstrate that simply mixing multiple datasets and training a single large end-to-end speech recognition model, called SpeechStew, can yield strong empirical results.
    \item We demonstrate the transfer learning capabilities of SpeechStew. When encountering a new dataset, one could simply fine-tune a generic pretrained model on unseen data to yield strong empirical results.
\end{enumerate}

\section{SpeechStew}

In this section, we describe the model and training data setup of SpeechStew. We also describe our transfer learning setup for fine-tuning on new unseen tasks. 

\subsection{Model}
In our implementation, SpeechStew uses the Conformer \cite{gulati-interspeech-2020} RNN-T \cite{graves-icml-2012} architecture. We experiment with both the 100M parameter \cite{gulati-interspeech-2020} and the 1B parameter configuration \cite{zhang-arxiv-2020}. We find that wav2vec pre-training \cite{baevski-arxiv-2020} is needed to train the 1B parameter model \cite{zhang-arxiv-2020}. We apply the default hyperparameters from prior work \cite{gulati-interspeech-2020,zhang-arxiv-2020} including the learning rate schedule. We do not incorporate an external language model.

\subsection{Multi-domain Training}
\label{sec:mdtraining}
We combine the following datasets without any form of re-weighting or resampling to construct the training set for SpeechStew:
\begin{enumerate}[noitemsep]
    \item AMI \cite{carletta-mlmi-2005}. AMI is approximately 100 hours of meeting recordings.
    \item Common Voice \cite{lrec-2020}. Common Voice is a crowd-sourced open licensed speech dataset. We use the version 5.1 (June 22 2020) snapshot with approximately 1500 hours. The data was collected at 48 KHz, and we resampled it to 16 KHz.
    \item English Broadcast News (LDC97S44, LDC97T22, LDC98S71, LDC98T28). English Broadcast News is approximately 50 hours of television news.
    \item LibriSpeech \cite{panayotov-icassp-2015}. LibriSpeech is approximately 960 hours of speech from audiobooks.
    \item Switchboard/Fisher (LDC2004T19, LDC2005T19, LDC2004S13, LDC2005S13, LDC97S62). Switchboard/Fisher is approximately 2000 hours of telephone conversations. The data was collected at 8 KHz, and we upsampled it to 16 KHz.
    \item TED-LIUM v3 \cite{rousseau-lrec-2012,hernandez-specom-2018}. TED-LIUM is approximately 450 hours of TED talks.
    \item Wall Street Journal (LDC93S6B, LDC94S13B). WSJ is approximately 80 hours of clean speech.
\end{enumerate}

\subsection{Transfer Learning}
We demonstrate the transfer learning capabilities of SpeechStew. Once we have a general purpose SpeechStew model (trained on the datasets mentioned in Section \ref{sec:mdtraining}), we can fine-tune and adapt SpeechStew onto a new task. CHiME-6 \cite{watanabe-arxiv-2020} is a noisy low resource dataset set, which contains approximately 40 hours of distant microphone conversational speech recognition in everyday home environments. CHiME-6 is difficult for end-to-end speech recognition models to train directly due to over-fitting issues \cite{andrusenko-arxiv-2020}. We fine-tune SpeechStew on CHiME-6 to demonstrate the transfer learning capabilities.

The transfer learning capabilities of SpeechStew are extremely practical. It implies we can train a general purpose model once, then fine-tune to specific low resource tasks. This can be done at a very low cost, since fine-tuning typically requires only a few thousand steps, compared to $\approx$100k steps needed to train a model from scratch.

\begin{table*}[t]
\centering
\resizebox{\textwidth}{!}{%
\begin{tabular}{lccccccccc}
\toprule
\bfseries Task&\multicolumn{2}{c}{AMI} & Common Voice & \multicolumn{2}{c}{LibriSpeech} & \multicolumn{2}{c}{Switchboard/Fisher} & Tedlium & WSJ \\
\midrule
 & IHM & SDM1 & & clean & other & SWBD & CH & & eval92 \\
\midrule
\bfseries Prior Work (no LM) \\
\quad Single domain & & & 16.9$^\dagger$ \cite{likhomanenko-arxiv-2020} & 1.5 \cite{zhang-arxiv-2020} & 2.7 \cite{zhang-arxiv-2020} & & & 7.5 \cite{likhomanenko-arxiv-2020} & 9.3 \cite{sabour-iclr-2019} \\
\quad Multi-domain & 12.2$^\ddagger$ \cite{kanda-arxiv-2021} & \textbf{21.2}$^\ddagger$ \cite{kanda-arxiv-2021} & 15.5$^\dagger$ \cite{likhomanenko-arxiv-2020} & 3.0 \cite{likhomanenko-arxiv-2020} & 7.3 \cite{likhomanenko-arxiv-2020} & 6.3 \cite{likhomanenko-arxiv-2020} & 10.7 \cite{likhomanenko-arxiv-2020} & 6.9 \cite{likhomanenko-arxiv-2020} & 3.4 \cite{likhomanenko-arxiv-2020} \\
\midrule
\bfseries Prior Work (with LM) \\
\quad Single domain & 17.5 \cite{sun-arxiv-2021} & 36.4 \cite{kanda-icassp-2019} & 13.6$^\dagger$ \cite{likhomanenko-arxiv-2020} & \textbf{1.4} \cite{zhang-arxiv-2020} & \textbf{2.6} \cite{zhang-arxiv-2020} & 4.9 \cite{wang-interspeech-2020} & 9.5 \cite{wang-interspeech-2020} & 5.6 \cite{zhou-icassp-2020} & 2.9 \cite{hadian-interspeech-2018} \\
\quad Multi-domain & & & 10.6$^\dagger$ \cite{likhomanenko-arxiv-2020} & 2.1 \cite{likhomanenko-arxiv-2020} & 4.4 \cite{likhomanenko-arxiv-2020} & 5.5 \cite{likhomanenko-arxiv-2020} & 9.1 \cite{likhomanenko-arxiv-2020} & \textbf{5.2} \cite{likhomanenko-arxiv-2020} & 2.0 \cite{likhomanenko-arxiv-2020} \\
\midrule
\bfseries Our Work (no LM) \\
\quad Single Domain Baseline (100M) & 26.1 & 40.5 & 16.3 (13.8$^\dagger$) & 2.1 & 4.4 & 5.6 & 9.7 & 7.6 & 28.2 \\
\quad SpeechStew (100M) & \textbf{9.0} & \textbf{21.7} & 12.1 (9.7$^\dagger$) & 2.0 & 4.0 & \textbf{4.7} & \textbf{8.3} & \textbf{5.3} & \textbf{1.3} \\
\quad SpeechStew (1B) & 9.5 & 22.7 & \textbf{10.8 (8.4$^\dagger$)} & \textbf{1.7} & \textbf{3.3} & 4.8 & 10.6 & 5.7 & \textbf{1.3} \\
\bottomrule
\end{tabular}}
\caption{Speech recognition word error rates (\%) across multiple tasks. SpeechStew achieves SoTA or near SoTA across many tasks. Our SpeechStew 1B model uses wav2vec pre-training on LibriLight. SpeechStew does not use a separate language model. $^\dagger$We follow \cite{likhomanenko-arxiv-2020} and remove punctuations during evaluation. $^\ddagger$Concurrent work \cite{kanda-arxiv-2021}.}
\label{table:results}
\end{table*}

\begin{table}[t]
\centering
\begin{tabular}{lcc}
\toprule
\bfseries Model & \bfseries Dev & \bfseries Eval \\
\midrule
\textbf{Prior Work (with LM)} \\
\quad Official HMM Baseline \cite{watanabe-arxiv-2020} & 51.8 & 51.3 \\
\quad HMM \cite{medennikov-chime-2020} & \textbf{36.9} & \textbf{38.6} \\
\quad RNN-T \cite{andrusenko-arxiv-2020} & 49.0 \\
\midrule
\textbf{Our Work (no LM)} \\
\quad\textbf{Zero-Shot (never seen CHiME-6)} \\
\quad\quad SpeechStew (100M) & 54.9 & 57.2 \\
\quad\quad SpeechStew (1B) & 39.2 & 53.7 \\
\quad\textbf{Fine-tuned with CHiME-6} \\
\quad\quad Baseline (100M) & 70.0 & 66.7 \\
\quad\quad SpeechStew + Fine-tune (100M) & 33.1 & 40.6 \\
\quad\quad SpeechStew + Fine-tune (1B) & \textbf{31.9} & \textbf{38.9} \\
\bottomrule
\end{tabular}
\caption{We apply transfer learning and fine-tune SpeechStew on CHiME-6.}
\label{tab:chime}
\end{table}

\section{Experiments}

As mentioned in Section \ref{sec:mdtraining}, we use AMI, Common Voice v5.1, English Broadcast News, LibriSpeech, Switchboard/Fisher, TED-LIUM v3, and Wall Street Journal to train our SpeechStew model. For AMI, we use both individual headset microphone (IHM) and single distant microphone (SDM1) data. We resampled the data to 16 KHz for Common Voice and Switchboard/Fisher (which were 48 KHz or 8 KHz respectively). We use 80 dimensional filter bank coefficients as our input features. We evaluate our model on several tasks: AMI (IHM and SDM1), Common Voice, LibriSpeech (clean and other), Switchboard (Switchboard and CallHome subsets of Hub5'00), Tedlium, and WSJ eval92. We score our results with Kaldi scripts \cite{povey-asru-2011}. For Common Voice, we evaluate with and without an extra step of text normalization by dropping punctuations to make our results comparable to \cite{likhomanenko-arxiv-2020}. We do not apply any other form of text normalization during training or evaluation.

We build single task mode baselines, where the models are trained only on their respective domains. We use the Conformer 100M architecture \cite{gulati-interspeech-2020} for these baselines; we found the 1B model to overfit and perform dramatically worse. We perform model selection via the development sets per baseline task.

Our SpeechStew model uses the 100M parameter and 1B parameter Conformer architecture \cite{gulati-interspeech-2020,zhang-arxiv-2020}. We used the default experimental settings of these references to train the models. We train all our SpeechStew models for exactly 100k steps, without any model selection.

The 100M parameter model, referred to as ``Conformer L" in \cite{gulati-interspeech-2020}, is trained for 100k steps using Adam optimization with $\beta_1=0.9$, $\beta_2=0.98$ and a transformer learning rate schedule (section 5.3 of \cite{vaswani-nips-2017}) with 10k warm-up steps and peak learning rate 2e-3. The batch size is set to 8192. We find that using a large batch is important for getting good results. Residual dropout with rate 0.1 is applied, as is weight-noise \cite{graves-nips-2011} and L2-regularization with weight 1e-6. Exponential-moving-averaged model weights with decay rate 0.9999 are used for evaluating the model. We apply adaptive SpecAugment \cite{park-interspeech-2019,park-icassp-2020} with two frequency masks of size parameter $F = 27$, and ten time masks with maximum time-mask ratio $p_S = 0.05$ to augment the input spectrogram.

For the 1B parameter model, ``Conformer XXL" of \cite{zhang-arxiv-2020}, we follow this reference to apply LibriLight wav2vec 2.0 \cite{baevski-arxiv-2020} pre-training to initialize the network. LibriLight \cite{kahn-arxiv-2019} is approximately 60k hours of unlabelled data, in-domain with LibriSpeech. As in \cite{zhang-arxiv-2020}, we found the wav2vec pre-training to be critical to optimize these large neural networks. Following \cite{baevski-arxiv-2020}, we use non-quantized pre-training with sampling probability 0.065 and mask size 10. The wav2vec contrastive loss is optimized via Adafactor optimization \cite{shazeer-icml-2018} with parameters $\beta_1=0.9$, $\beta_2=0.98$. We use a transformer learning rate schedule for pre-training with peak learning rate 2e-3 and 25k warm-up steps. Pre-training the encoder of the model for 400k steps, we carry out supervised training on SpeechStew for 100k steps with batch size 2048. The encoder and decoder are optimized separately with two Adafactor optimizers (with the same beta parameters as in pre-training) and two transformer learning rate schedules. Peak learning rates 3e-4/1e-3 and warm-up steps 5k/1.5k are used for the encoder/decoder schedules, respectively. The same SpecAugment policy as the 100M model is used to train this model, while exponential moving averages are not used.

Table \ref{table:results} summarizes our results. We emphasize that the reported performance of SpeechStew is obtained without the use of an external language model.

SpeechStew 100M outperforms almost all prior work, including those using strong language models. On AMI-IHM and AMI-SDM1, SpeechStew obtains 9.0 WER and 21.7 WER respectively. On Common Voice, SpeechStew obtains 12.1 WER or 9.7 WER with punctuation normalization. On LibriSpeech, SpeechStew obtains 2.0 and 4.0 on the clean and other test splits. These results are worse than \cite{zhang-arxiv-2020}, where noisy student training \cite{park-interspeech-2020} utilizing extra unlabelled data (LibriLight) has been applied.
On Switchboard/Fisher, SpeechStew obtains 4.7 WER on the Switchboard split, and 8.3 WER on CallHome split. On TED-LIUM v3, SpeechStew obtains 5.3 WER, slightly behind the 5.2 WER of \cite{likhomanenko-arxiv-2020}, which has been obtained with the use of a language model. On WSJ, SpeechStew obtains 1.3 WER. To the best of our knowledge, our SpeechStew model outperforms all prior work on AMI, Common Voice, Switchboard and WSJ. Overall, we find SpeechStew to achieve SoTA or near-SoTA across a wide variety of tasks.

SpeechStew 1B with LibriLight wav2vec pre-training \cite{baevski-arxiv-2020} exhibits improved performance on Common Voice and LibriSpeech compared to the 100M model, while the other tasks suffer from a small relative degradation. One possible explanation for this phenomenon is that the LibriLight pre-training data used to initialize the network is in-domain with the LibriSpeech task, suggesting that the unsupervised pre-training data may bias the network toward a certain domain. We, however, also find that the 1B model exhibits better transfer learning capabilities in the next section.

\subsection{Transfer Learning and CHiME-6}

CHiME-6 \cite{watanabe-arxiv-2020} is a low resource noisy speech dataset. It is especially challenging due to the small size and noisy audio conditions. We perform front end enhancement following the official recipe \cite{watanabe-arxiv-2020}. We use BeamformIt to enhance our front end during training. We use guided source separation \cite{boeddecker-chime-2018} with 12 channels to enhance our front end for evaluation.

We fine-tune our 100M and 1B SpeechStew models with the CHiME-6 data, and compare these results against baselines obtained by training a 100M parameter Conformer model and a 1B-parameter Conformer model (with LibriLight-only pre-training) with CHiME-6. Table \ref{tab:chime} summarizes our results. We have not recorded results for the 1B-parameter baseline model, as we have failed to train the model to exhibit non-trivial performance. We have adjusted the dropout rate, the weight-decay parameter, augmentation parameters, learning rate and warm-up steps with respect to the default values from \cite{gulati-interspeech-2020} and \cite{zhang-arxiv-2020} to optimize the dev-set WER.

For the 100M models, we apply a universal dropout rate for the input, residual connections for the feed-forward, convolutional and attention units. For the baseline model, a universal dropout rate of 0.5 is applied, with L2-regularizer weight 1e-4. We only use two time masks with parameters $T=80$ and maximum time-mask ratio $p_S=0.25$ for augmenting the input. The default values for the learning rate schedule (2e-3 peak learning rate with 10k warm-up steps) are used. When fine-tuning from SpeechStew, a universal dropout rate of 0.4 and L2-regularizer weight 1e-6 is used. The SpecAugment parameters are set to be equivalent to the baseline, while the peak learning rate and warm-up steps for the learning rate schedules are set to 1e-4 and 2k steps, respectively. The best dev-set performance when fine-tuning from SpeechStew is achieved at 4k steps, compared to 60k steps for the baseline.

When fine-tuning the 1B SpeechStew model, L2-regularization is turned off. SpecAugment is applied with three frequency masks with $F=10$ and eight time masks with $T=96$ and $p_S=0.08$. Both the encoder and decoder learning rate schedule parameters are set to have peak learning rate 2.5e-4 and 8k warm-up steps. Best dev-set performance is achieved upon 6k steps of training.

The official CHiME-6 HMM baseline \cite{watanabe-arxiv-2020} achieves 51.8 and 51.3 WER on the dev and eval set respectively. A strong HMM model with a strong language model \cite{medennikov-chime-2020} achieves 36.9 and 38.6 WER on the dev and eval set respectively. The previous best end-to-end model \cite{andrusenko-arxiv-2020} achieves 49.0 on the dev set. The SpeechStew 1B parameter model in the zero-shot setting (having never seen any CHiME-6 data before), achieves 39.2 and 53.7 WER on the dev and eval set respectively. Further fine-tuning our SpeechStew model on the CHiME-6 data results in 31.9 and 38.9 WER on the dev and eval set respectively. We note that our results, obtained without using any external language model, compares to prior work that have utilized strong language models. This demonstrates that the SpeechStew model already possesses a significant amount of knowledge that is useful for the CHiME-6 task.

We make two observations regarding the effectiveness of SpeechStew on ChiME-6 in particular. The first is that the amount of training time spent on the low-resource CHiME-6 data is greatly reduced. This is a natural product of the fine-tuning process. However, SpeechStew's construction allows us to skip to that stage of the training process, thus reducing the risk of overfitting. Secondly, SpeechStew appears to be robust enough to handle noisy data. SpeechStew spends more time learning the salient features from a larger pool of less noisy data, and less time working with the more noisy CHiME-6 data.

\input{relatedwork.tex}
\input{discussion.tex}

\section{Acknowledgements}

We give thanks to Chung-Cheng Chiu, David Fleet, Naoyuki Kanda, Bo Li, and Samy Bengio for their  discussions and review of the manuscript.

\bibliographystyle{IEEEtran}

\bibliography{paper}

\end{document}

%% file: relatedwork.tex
\section{Related Work}
Mixing multiple datasets together to train one neural network is not new. Kaldi's \texttt{multi\_en} recipe \cite{kaldi-multi-en} has a very similar setup to ours. Narayanan et al.~\cite{narayanan-slt-2018} mixed several proprietary commercial datasets, totally more than 160k hours of training data to train one neural network as well. Likhomanenko et al.~\cite{likhomanenko-arxiv-2020} also mixed several datasets together to train end-to-end speech recognition models.~\cite{chojnacka-arxiv-2021} mixed several languages together to train a speaker verification system. One of the key difference between our work and prior work is the model size, we scale to much larger models.

Neural networks often benefit from scaling up the number of parameters in practice~\cite{nakkiran-iclr-2020}. This has been observed in language modelling \cite{devlin-naacl-2019,kaplan-arxiv-2020,brown-neurips-2020}, computer vision \cite{he-cvpr-2016,kaplan-arxiv-2020,chen-neurips-2020}, and speech recognition \cite{dean-nips-2012,zhang-arxiv-2020}. Similarly, SpeechStew leverages large 100M and 1B parameter models to yield strong empirical results. However, our model sizes still pale in comparison to those found in the language modelling literature, for example 175B parameters in GPT-3 \cite{brown-neurips-2020}.

Transfer learning is a popular technique found in computer vision \cite{jia-arxiv-2014,kornblith-cvpr-2019,kolesnikov-arxiv-2019} and natural language processing \cite{devlin-naacl-2019,brown-neurips-2020}. A general purpose model, which is trained either on a large corpus of unlabeled data or labeled data, is then finetuned on a low resource task. SpeechStew also adopts the pretrain, finetune setup for transfer learning. We train SpeechStew on a large labeled corpus, and finetune on CHiME-6.
There has also been prior work on pretraining on unlabelled audio, and finetuning on supervised data \cite{schneider-arxiv-2019,baevski-arxiv-2020}. Our work is focused on supervised pretraining and supervised finetuning. However, we still leverage unsupervised wav2vec pretraining in our 1B parameter models. We believe these techniques may be additive, especially as one scales to very large models. Finally, the concurrent work of \cite{kanda-arxiv-2021} also applied supervised pretraining and finetuning; they pre-trained on 75k hours of proprietary labelled data, and finetuned on AMI to achieve very strong empricial results.

%% file: discussion.tex
\section{Discussion}
        
Deep learning models have made significant progress from two simple principles:
\begin{enumerate}[noitemsep]
    \item Train on more data \cite{krizhevsky-nips-2012,brown-neurips-2020}.
    \item Train larger and deeper neural networks \cite{he-cvpr-2016,devlin-naacl-2019}.
\end{enumerate}

Supervised data is expensive to acquire. SpeechStew takles this problem by simply mixing all publicly available speech recognition data. Our approach leverages on currently available resources, labelled and unlabelled. We hope our work will encourage future research to leverage on all training data available, as opposed to training on only task specific datasets.

Training large models is expensive, and impractical to do frequently (especially when one regularly encounters new tasks or new data). Our work on transfer learning demonstrates that one can simply finetune a pretrained model for only a few thousand gradient steps and achieve strong results. This is inexpensive and very practical. We hope our work will encourage further research to leverage on transfer learning in speech recognition.